# Hybrid Reasoning and the Future of Iconic Representations

Catherine RECANATI

*LIPN – CNRS UMR 7030, Université Paris 13*
*Institut Galilée, Av. J-B. Clément, 93430 Villetaneuse, France*
catherine.recanati@lipn.univ-paris13.fr

**Abstract**. We give a brief overview of the main characteristics of diagrammatic reasoning, analyze a case of human reasoning in a mastermind game, and explain why hybrid representation systems (HRS) are particularly attractive and promising for AGI and Computer Science in general.

**Keywords**. Diagrammatic representation. Iconic representation, Analogical representation. Hybrid representation systems. Cognitive modeling and reasoning.

**Introduction**

Logical linguistic representations have a high power of abstraction and many people think that they can model our reasoning abilities, partly because our knowledge expresses in linguistic terms, and also because our formal tools are built on alphanumerical representations. Nevertheless, inferential systems solely based on textual representations are very inefficient. Moreover, these systems raise difficulties at the representational level, because they require a complete specification of the concrete and abstract properties of the modeled objects. This is why computer scientists, used to think in terms of data structures, have early defended the use of diagrammatic representations, for instance in problem solving, on the basis of the fact that these representations were better adapted to specific domains (see [1] for an historical survey and critiques of logicist AI).

Although commonly used in Science, for instance in Mathematics or Physics, diagrammatic representations have long suffered from their reputation as mere tools in the search for solutions. At the beginning of the 90's, Barwise and Etchemendy (B&E) have strongly denounced this general prejudice against diagrams ([2], [3], [4]). To cope with complex situations, they defended a general theory of valid inferences that is independent of the mode of representation, and these works lead on the first demonstration that diagrammatic systems can be sound and complete [5].

As far as human reasoning is concerned, there are many examples using non linguistic form of representation, and, to quote B&E, "human languages are infinitely richer and more subtle than the formal languages for which we have anything like a complete account of inference. [...]. As the computer gives us ever richer tools for representing information, we must begin to study the logical aspects of reasoning that uses nonlinguistic forms of representation" [2].

Following in B&E footsteps, our general project is to defend the interest of hybrid representation systems (HRS) – i.e. systems linking together several kinds of representations. We claimed in [6] and [7], that only HRS could yield to the building of models of reasoning, both computationally efficient and cognitively plausible.

In this paper, we will first recall the most interesting characteristics of diagrammatic inferential systems, and add some comments about an example of human hybrid reasoning in a mastermind game. In the next section, we will give some arguments for the systematic study (and use) of HRS in AGI and cognition modeling, and some hints for their usefulness in program specification and semantics.

## 1. Some characteristics of diagrammatic inferential systems

In [2], B&E emphasized that the main properties of diagrammatic systems derive from the existence of a syntactical homomorphism between icons and represented objects. In many cases, this homomorphism yields to a very strong property called *closure under constraints*. In closed under constraints systems, the consequences of initials facts are included de facto in the representation and do not require extra computation. This makes these systems very efficient. As we have underlined in [6] and [7], this also shows a deep duality between two modes of reasoning.

Linguistic (or traditional logical) reasoning requires: (1) the representation of initial properties of objects; (2) an explicit representation of abstract properties (or relations among objects); and (3) a computational mechanism linking the two sources of information (to establish the validity of a non-explicit consequence). Thus, by construction, such systems require calculations. For instance, if you know that Ann is on the left of Gaston on a bench, and that Gaston is on the left of Isabel, you need to add that the relation "be on the left of" is transitive to prove that Ann is on the left of Isabel.

To the opposite, diagrammatic reasoning usually does not require the explicit representation of such abstract properties, because these properties are taken automatically into account by syntactic constraints on the representation itself. In our example, an iconic representation of the first fact will look like the (left) juxtaposition of two symbols (say, A for Ann and G for Gaston, as in: A G); and the second fact will yield to the juxtaposition of a third symbol (say, I for Isabel), as in:   A   G    I.

Thus, you will just "see" on the resulting representation that A is on the left of I, without any computation. Since many consequences automatically appear on representations, diagrammatic systems provide an easy treatment of conjunctions and are computationally very efficient. Unfortunately, they have difficulties with disjunctive cases[i]. Alternatives may require the use of several diagrams, which must then be traversed one after the other, as in the linguistic case[1]. Note also that in many diagrammatic systems, each representation corresponds to a genuine situation, and that contradiction is impossible to represent (which can be good or bad depending on what you need to represent).

Many researchers have tried (in the nineties) to analyze diagrammatic inferential systems properties and closure under constraints in particular. For Stenning and Oberlander (S&O) [9], diagrammatic representations seem mainly to differ from linguistic ones by a more limited power of abstraction, but greater computational efficiency. They claimed that there are three classes of representational systems: the MARS (Minimal Abstraction Representational Systems), the LARS (Limited Abstraction Representational Systems) and the UARS (Unlimited Abstraction Representational Systems). They argue that this hierarchy of representational systems is analogous to that of languages isolated by Chomsky, and that most diagrammatic representation systems are LARS. A MARS is a system in which a representation corresponds to a unique model of the world under the considered interpretation. For instance in a mastermind game, a row of letters standing for a row of colored pawns, as [B B Y Y R], will be a minimal abstraction representation of a possible solution. However, you can easily augment the number of models captured in a MARS by introducing new symbols that allow abstracting on representations. For instance, in the mastermind example, you can have a "-" symbol standing for an undetermined color, as in [B B - Y R]. Such systems can quantify massively on possible models, but cannot specify arbitrarily complex dependences between the specified dimensions. This is why S&O called them LARS. They claimed that only linguistic symbols, added to a representation, could allow the description of arbitrarily fine dependences between dimensions. They defined a LARS as "a system that keeps its representations simple, and keeps assertions out of its keys" and claimed that most diagrammatic inferential systems are LARS.

---

[1] The difficulty with disjunction reinforces the thesis that cognitive representations are mainly diagrammatical, because human performances are better in conjunctive than in disjunctive cases [8].

S&O identify the restricted capacity of diagrammatic systems with a property called "specificity", which requires information of a certain kind to be explicit in all interpretable representation. In [10], Perry and Macken (P&M) have opposed to this strong notion of specificity (i.e. the mandatory specification of values of properties other than the one you try to represent) the notion of "determined character" due to Berkeley. Berkeley's notion of a determined character is that it is not possible to represent an object as having a certain property, without representing at the same time a specified value for this property. Thus, I cannot represent a triangle on a figure, without ending with a particular triangle. As well, it is not possible to represent a colored object on a drawing without specifying its color, but I can perfectly say, « this object has an interesting color », without specifying which one[ii]. For P&M, closed under constraints systems have, in addition to this determined character, a property called "localization" (already identified by Larkin and Simon in [12]). Localization is more important than specificity to characterize diagrammatic representations. Nevertheless, there are two properties of localization. The one identified by P&M is a purely logical property also called *unique token constraint*. It is the property of using only one token of a symbol to represent an object. This property disappears generally when you use a typed system[iii]. Finally, P&M distinguish five kinds of representation going from text to images: graphic texts, charts, diagrams, maps and pictures. Their categorization uses two additional properties, iconicity and a constraint and systematic homomorphism (required to handle closure under constraints).

As far as geometric or spatial aspects are concerned, Macken, Perry, and Hass emphasized the importance of *iconicity* in [13]. Iconicity allows representations with richly grounded meaning – that is meaning whose relation to form is not arbitrary. An iconic sign may have a readily inferable meaning (RIM), an easily remembered meaning (ERM), or an internally modifiable meaning (IMM). Road signs provide numerous examples of ERM, RIM and IMM (for instance, signposting bends). There also are many examples of symbols having a RIM in musical scores (as for instance, crescendo situated under the stave). However, iconicity is only partially analyzed until now, and IMM is still puzzling. We think that it could be sometimes linked to the syntactic homomorphism, because our personal conclusion is that the main distinction between linguistic (or symbolic) representation systems and analogical representation systems (as diagrammatic systems) must be characterized in terms of the power of the meta-language required to provide the semantics of the system. In the analogical case, the metalanguage needs to reference syntactical properties of the object language, while in the symbolic case, this is not obligatory[iv].

## 2. Hybrid human reasoning in mastermind

The preceding section recalls that iconic representations can be first class citizen, i.e. valid syntactical objects in inferential systems. It also underlines what iconic representations are good for and what they are not. At first sight, a limited power of abstraction and the request of a unique syntactical homomorphism are restrictive, and situations to which purely diagrammatic reasoning applies seem limited[2]. Nevertheless, graphical and textual representation systems being complementary (at representational and algorithmic levels), the shortcomings of both systems can disappear in HRS. Therefore, the preceding review acts as a critique of current approaches to reasoning, which tend to emphasize only one mode, diagrammatic or linguistic, and are set up in opposition to the other mode (e.g. the mental logic vs. mental models debate). Let us now look at an example of hybrid human reasoning in a mastermind game.

Mastermind[3] is well suited to the study of human reasoning, because it constrains the player to perform logical reasoning. Furthermore, the geometry of the grid encourages the players to use diagrammatic representations. For most players, reasoning is fragmented and opportunistic, and consists in partial deductions using several types of representations. In [14], we highlighted this hybrid character: most of deductions are graphical, while the model under construction partially expresses verbally. In fact, the use of graphical representations mitigates limitations in the cognitive capacities of the player, anchoring reasoning on inexpensive visual capacities, and relieving thus verbal memory. In return, visual capacities being themselves restricted, the shape of the diagrams and the ordering of hypotheses are biased (this because, even when they express verbally, hypotheses are also grounded on the grid). For instance, the left-to-right order (of pins and pawns) and the ease of visual translations, influence the choice of hypotheses to be considered first. Nevertheless, some players use these biases to develop their own strategy of resolution in an intelligent way.

We have insufficient room here to report all of our observations, but we can shortly comment a game of one player (grid on Figure 1). The grid ensures the memorizing of preceding results, but, as we will see, it is also a geometrical support for organizing proof and backtracking. Our player separates her game in two phases: first determining the colors, and then determining the places. In both phases, she uses representations that can be qualified as mental models because they are very similar to those of Johnson-Laird [15]. The interesting fact here is that these models (which also correspond to LARS of S&O) are ordered both by increasing order of specificity, and by decreasing order of probability. This makes backtracking easier, since the model considered next is determined, and guarantees a quick convergence to the solution, since these models are in decreasing order of probability.

---

[2] Contrary to what may seem initially, graphical representations are not only helpful in modeling situations where a (concrete) spatial homomorphism applies.

[3] The game consists in discovering a hidden row of five colored pawns. One player (the leader) hides a configuration of pawns. The second player can then dispose on a grid a tentative configuration of pawns, and the leader replies by posting pins (on the right) indicating if and how pawns correspond to the solution one's. A white pin means a good position and color for one pawn, and a black one a misplaced color. The rows remain visible during the game, and the player has to find out the solution with a limited number of rows.

|   |          |        |
|---|----------|--------|
| 6.| R R G Y G| o o o o o |
| 5.| G R R Y G| o o o ● ● |
| 4.| R G R Y G| o o o ● ● |
| 3.| R R R G G| o o o ● |
| 2.| O O B B B|        |
| 1.| B B Y Y R| o ●    |

**Figure 1.** Game of an experienced player

The player begins on row 1 by her favorite attempt (a 2/2/1 distribution), which possible replies revealed being statistically more informative than those of other colors distributions (such as 3/2, 4/1, 5, 1/1/3 or 1/1/1/2, etc.). Given the pins on the right side, she considers first the interpretation displayed on Figure 2, i.e. that one blue is placed correctly, one yellow misplaced, and that there is no red. (She might take in his hand a blue and a yellow pawn to help memorizing, and note mentally that the three colors are exhausted).

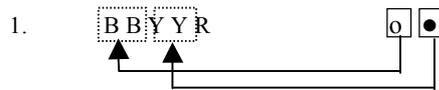

**Figure 2.** A first interpretation schema

We note this mental model by [1B] [1Y] (and "no red") – using square brackets for the notion of exhaustion introduced by Johnson-Laird. (Note however that the model behind the schema of Figure 2 is more specific, since it includes some information on places, but in this first phase of the game, the player does not pay much attention to them). Then, she plays the second row, trying new places for blue (anticipation on future reasoning about blue places), and introducing a new color: orange. By luck, both orange and blue are missing colors, and the interpretation of the second row is obvious. Blue being excluded, she switches to a new model based on a new interpretation of the first row: [1 Y] 1R.

Then, she plays the third row both to try new places for red, and to try a new color. Getting four pins as a result, she concludes easily that the colors of the solution must be yellow, red and green. Given that there is only one yellow, she considers first [1Y] [2R] [2G] (which seems more probable than [1Y] [3R] [1G]). She then begins reasoning on places and supposes that on the first row, it is the first left yellow that is correct (we will note this model by [– – Y – –], knowing that the empty places must be filled by the missing pawns within [1Y, 2R, 2G]).

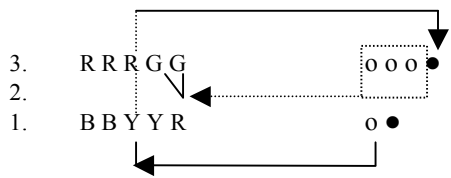

**Figure 3.** A diagrammatic reasoning

With the diagrammatic reasoning illustrated in Figure 3 (start following the arrows from the first row), she infers that on row 3, a red is misplaced, and thus, two greens well placed ([– – Y G G]). The solution should be [R R Y G G], but this conflicts with the four pins of row 3, which should then be all white. Thus, she has to backtrack and reconsider the position of the yellow pawn on the first row ([– – – Y –]).

A graphic reasoning very similar to the preceding one reveals that in this case, the left green is misplaced and the right one correct (i.e. [– – – Y G]). She then tries a fourth plausible row, but is this time unlucky. Nevertheless, colors are confirmed and she knows by experience that, getting 3 white and 2 black pins means that two pawns have just to be exchanged to give the solution. The two pawns to switch are to be found in the first three pawns [R G R – –], thus the green must be exchanged with one of the two reds. She tries [G R R Y G] on row 5, but is unlucky again. However, there is now only one solution for the switch, and she wins on the last row.

An interesting fact about this game is the use of graphical inferences as those depicted on Figure 3. There are other sorts of graphical inferences used by experienced players. For instance, by focusing on the common parts of several rows, inferences can be draw from the requested mappings between the set of common pins and the set of common pawns. All of these inferences are in a way "local" within the global reasoning, and they use creative graphic schemas mapped on the fly onto the grid. At a higher level, the strategy used by this player consists in a systematic ordering of the possibilities opened by a given row. This kind of strategy is often used. At the beginning, the reasoning is rooted on the first row, and the most probable model is considered first, here with a left-to-right bias in case of equality. For instance, in the preceding game, the ordering of the several models compatible with the first row (without considering places) is the following:

[B][Y] no red  < [B] R no yellow < [Y] R no blue.

The players compare competitive models and their relative probabilities directly from the number of pins and pawns. This is why some players have a tendency to prefer continuous color arrangements to separate ones, because quantities (or mass) are then more salient, and the comparison analogically performed easier. In many cases, the player builds a model in easy stages by covering a lattice where models fit into each other on a branch (by being more specific). The nature of considered models is not always as systematic as in our example, and may vary among players and/or situations. Nevertheless, an important fact is that these models layout on the grid in a visual manner. Figure 5 gives examples of several models (fitting together graphically).

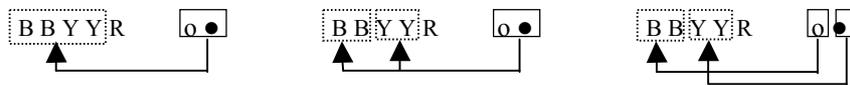

**Figure 5.** Some graphical schemas of interpretation

However, concerning our player, the global reasoning path is oriented by two directions (both grounded on the grid): (1) a left-to-right orientation of the possible models within a row, and (2) the natural vertical ordering of the rows. This systematic ordering helps remembering which model has to be consider next in case of backtrack. This global strategy applies as well in the second phase of the game. Here for instance, the ordering on the first row is:

[– – Y – –] < [– – – Y –] < [– – – – R]

The first model [– – Y – –] was quickly eliminated, and [– – – Y –] evolved progressively in a more specific solution.

Another interesting fact about this example is that diagrammatic representations prevent here from incoherence, instead of introducing errors (as many people claimed they merely do). Here this is due to the use of limited abstraction diagrams in which contradiction is impossible to represent. Furthermore, partially because of the specificity property mentioned in the first section, LARS appear to be good candidates for ordering models by inclusion. Models may also be orderly among other dimensions, by using probabilities or other specific attributes.

From this point of view, our example can be seen as a prototype for a family of programs, where information arises incrementally (here on each new row) and which are more or less determining (or approximating) the "solution" – thought of as a matrix of values. In such cases, the articulation of local (possibly graphical) subsystems within the lattice and the general level controlling flow (in charge of backtracking) is simple, because the role of each module is well definite. Each new information may bring specific constraints between specific values, and expresses partially in some subsystem, but the general program cannot be prepare to all of them. Then, other local heuristics or strategies will help and give a core to the general reasoning.

For instance, this sort of architecture could naturally apply to natural language processing, because text appears sequentially, both at discourse level and at sentence level. Suppose we have to process some text (i.e. already organized in words, but a similar architecture would apply to speech). Each word arrives with new information about the "meaning" of a sentence. Such meaning could locally be a matrix of several sorts of attributes (which might be values or actions) depending on what the software is supposed to do. With current semantics theories, it could be made of linguistic features from several domains (morphology, syntax, semantics, etc.). In each domain, there are specific constraints that can be handle by modular and more or less independent subsystems (for instance, in events semantics, you might have specific representations for time, space, causality, etc.). Thus, the general program may used statistics, proper strategies (as try to discover syntactical features first) or a left-to-right bias as in our mastermind example, to find a way through the several possibilities of filling up the mandatory features – without presupposing that some of these features (syntactic ones in particular) have to be completely determined first. Particular features may also be let undetermined, to keep the natural lack of precision in language.

## 3. Perspectives for Hybrid Representation Systems in AGI

HRS may lead to amazing results concerning efficiency. A paradox is that a given demonstration may be limited by a minimal cost in any symbolic system, and still be less costly in a hybrid system including and binding the two sorts of representations (iconic or symbolic ones). Note that there is nothing sophistic here, because in a hybrid system there is no need of a global language to bind its subsystems[4] (remind Gödel's proof). Furthermore, the articulation of several subsystems in a complex representational system, bases sometimes simply on the fact that they denote the same objects in the world, and therefore coherence between two subsystems has not necessarily to be handle. In the domain of reasoning, the objection that situations in which a unique homomorphism applies are rare is as well not too serious, because you can use several homomorphisms. The situation is just that the subsystems denote different properties of models or objects, and what expresses in one subsystem do not express necessarily in the other. Nevertheless, some information can be transfer from one system to another (on the basis of safe correspondences), endowing the global system with superior inferential and computational capacities. And there is no special need of an intermediate language.

---

[4] This is why B&E gave a theoretical justification of the two main algorithms implemented in *Hyperproof* [16] based on purely mathematical grounds, without using any intermediate language.

Contrary to what may seem initially, graphical representations are not only helpful in modeling situations where a spatial homomorphism applies. Their increased use in science is also due to their obvious ability to convey abstract meanings. Via space, they bring new possibilities of structuring and abstracting (compared to sequences of letters alone). From this point of view, HRS are definitively on top of traditional UARS in the hierarchy of S&O. Besides their application to reasoning, their systematic study should improve formalization in many domains, which are relevant for AGI, as cognitive science, natural language semantics and linguistics. There are many domains in semantics where iconic representations seem better suited than logical formalisms. In linguistics, the numerous schemas, found in works on time and aspect (for instance [17], [18], [19]), are an indication of the plausibility of this thesis. We believe that concerning these domains, it is due to the nature of our cognitive apparatus (see next subsection). We also believe that the addition of iconic features in theoretical languages or tools could bring major advances in other fields of Computer Science, less concerned by world representations, as for instance, in the domain of semantics of programming languages, or in software design in general. By way of conclusion, we add two subsections to reinforce these claims. The following are surely controversial proposals. (They are also rather independent and some might be valid, others not.)

*3.1. Outline of a model of the human mind (relation between thought and language)*

To further understanding of the human mind, its higher cognitive capacities, and more specifically the nature of the relation between language and thought, the goal is to develop a model of language understanding and use that attains observational adequacy, i. e. that is able to pass the Turing test. To achieve this goal, we must aim higher, by trying to reach explanatory adequacy, that is, to develop a model of how the system can reasonably acquire the "knowledge" (i. e., systems of knowledge/belief, etc.) that enables it to attain observational adequacy.

The only way a mind can acquire the rich variety of knowledge humans do acquire is to start with a strong innate basis. The only way to build a system with a strong innate basis is to organize this basis into modules that are well adapted to representing the aspects of the world they represent. This is because of the way the world is (it is rich and varied, and the basic conceptual apparatus needed to represent time and temporal relations, for instance, must use different resources obeying different constraints than that needed to represent spatial relations, or interpersonal relations and other minds, or causal interactions, etc). There are probably also general computational constraints (problems of tractability and expressive adequacy), and the need for revision within relevant constraints (as well as many other factors), which will determine the emergence of a set of modules.

The mind's rich set of innate modules, its "knowledge" about the world (including itself) is thus in the form of representational capacities. While it can be heuristically useful to formulate knowledge/beliefs about time, for instance, as a set of axioms (i. e., declaratively) it is more plausible to consider that the mind embodies this knowledge as a capacity for representation (for instance, for representing temporal entities and relations among them). The knowledge is then embedded as constraints on what can be represented, and it will be useful to approach the problem of specifying knowledge in a certain domain, as the problem of specifying a 'grammar' of possible representations in that domain (e. g. possible representations of temporal relations among situations — precedence, overlapping, inclusion).

Besides this rich set of domain-specific modules, the mind needs to be equipped with a set of procedures for developing and enhancing the innate basis. While some of these are no doubt domain-specific, others must be domain-independent. We hypothesize that the human mind starts life with an innate basis for domain-specific knowledge that is more analogical or diagrammatic in nature, and that one of the important ways it develops is in the enrichment of the innate representational capacities with more symbolic representational capacities[5].

A mind that has the ability to choose how it will represent a particular problem it needs to solve, choosing from a repertoire of representational capacities that include more analogical and more symbolic notations is more flexible, hence more "intelligent" (more apt to solve its problems, hence to survive). We postulate that humans have this kind of mind. To handle this ability to choose between several representational capacities, and to keep its repertoire relatively unchanged (after a certain level of development), a mind needs also to have generic and global cognitive procedures to construct representations on the fly.

Following the general framework of cognitive approaches to language, we believe that linguistic forms are (partial and undetermined) instructions for constructing interconnected domains with internal structure. As claimed in [20] by G. Fauconnier, this construction takes place at a cognitive level C. This level is distinct from language structure. Constructions at level C are not "meanings", neither representations associated with any particular set of linguistic expressions. They are not representations of the world, or of models of world, or whatsoever of this sort. However, these constructions relate language to real world, and they provide various real-world inferences. They also are novel and different for each case of language use, and mental spaces and connections build up as discourse unfolds. The primary goal of (and primary evidence for) the approach in terms of interconnected domains is scientific generalization.

The first developments of Fauconnier "mental spaces" theory focus on processes of transfer from a source (or base) to a target. The capacity of organisms to carry out such projections lies at the heart of cognition in its many forms. The analyses given by Fauconnier are numerous and based on a rich array of linguistic data (counterfactuals; time, tense, and mood; opacity; metaphor; fictive motion; grammatical constructions; and quantification over cognitive domains). Further developments of the theory study another very interesting operation, conceptual blending [21], which also depends centrally on structure projection and dynamic simulation. Like standard analogical mapping, blending aligns two partial structures, but in addition, blending projects selectively to form a third structure. (Creativity in Science is often based on conceptual blending).

---

[5] Similar hypotheses relative to the architecture of mind and compatible with data (in psychology of reasoning), conclude to the existence of a meta-representational reflective level (i.e. handling meta-representation, as thought about thought, and the like) where slow logical inferences are drawn consciously.

All these works in Cognitive Semantics give us guidelines and examples to investigate in details how symbolic and iconic representations might relate in an intelligent complex system.

*3.2. Additional remarks from a Computer Scientist point of view*

The problem of the building (and, at first, of the description) of complex program architectures on computers is the concern of software engineering. We think that the numerous difficulties arising at this level are due to the deficiencies of our programming languages, in particular because they do not incorporate a more sophisticated level of description of the import features from other modules. Our claim is that they do not describe their own architecture (and therefore cannot incorporate import features at this level of description). We will not develop this claim here, although diagrams are obviously helpful for the description of architectures. We will only add a few remarks on iconicity (or on the non arbitrary shape of a symbol), to show that this dimension could be helpful at various levels of semantic description.

A first remark is that a non arbitrary shape character may appear in small touches, at the level of an isolated symbol, without even being included in a true iconic system (with proper analogical properties). For instance, a simple difference in the character font, as the addition of bold face, could modify traditional symbolic representations in a creative way. You can keep the old meaning for the new expression (for instance a value 0 and a value **0** both referring to zero as usual) and nevertheless have a supplementary meaning, relative to another dimension in the modeled world, or in the calculation process itself. You can for instance distinguish between a true (and final) value, from one that could still change, or be set by default by the system. Or, when added to more abstract symbols, as those describing the rewriting rules of a logic system, it could introduces second order rewriting rules, allowing to ignore intermediate terms (for instance, if not in boldface). Thus, traditional elimination rules could apply in a more efficient way (i.e. between distant elements), just by means of an additional graphical feature, defined and used at the level of the meta-language itself.

Our second remark is that, in the context of a computer, the general schema for the implementation of the homomorphism between syntactic representations and semantic representations do not stand on a simple line, as philosophers and logicians consider. It will translate into a program that will calculate, from internal representations (coming from our syntactic representations, by an operation of "internalization"), other internal representations – which we have to "externalize" if we want to get them explicit. Therefore, there are many other means to establish correspondences, or exploiting particular diagrammatic features, between all of these representations. In particular, some iconic relations may bind the syntax of the programming language used to that of the internal representations used, yielding to internally modifiable meaning (IMM). In reflective interpreters (cf. lisp), an object interprets as program or data, depending on its context of use. In such framework, the traditional data/program distinction vanished (as in machine languages). With reflective features, evaluation can be suspended or delayed (some functional languages implements lazy evaluation). Some programming languages may also have other specificities, as for instance, a pattern-matching operation as in PLASMA (an actor language of the eighties). Therefore, on a computer, very complex relations between the represented world and the representing world are virtually possible.

Another remark relative to the use of bold (or other such features) is that it can obviously be use to handle some notion of focus. Focus theories have not yet been successfully design, but it is a lack in our theoretical tools. There are many fields where some notion of focus would be of great help (in perception theory, in discourse theory, etc.). One reason of this failure might be precisely that the theories of focus require references to the underlying computational mechanism (as reflective properties of the programming language)$^v$.

If we take seriously the assumption of endnote IV, i.e., that the meta-language required to provide the semantics of a system has to reflect (in some way) the possibilities of configurations of terms in the representational language, then we have to investigate the following questions: what syntax do we need to easily provide the semantics of HRS? Would it be enough to add simple reflective and local graphical feature (as those of some of our programming languages) to a traditional functional and symbolic language, or should this syntax be trickier?

**Conclusion**

Works done so far on diagrammatic reasoning provide fragments of evidence about how people use iconic representations, and identify some of the problems raised by the project of AGI. Yet, there is still much to do to understand the variety of forms in which information can stored and manipulated in intelligent control systems. We believe that we could make important progress in studying in details the relation between iconic and symbolic features in hybrid representation systems, as well as in paying attention to them in the theoretical tools and symbolic languages that we use.

**Endnotes**

$^i$ However, contrary to what many authors have said, it is not difficult to represent disjunctive cases on diagrams, and we will see some exemplars in the next section (see Figure 5). It is also possible to have iconic symbols of second order in purely diagrammatic systems. C.S. Peirce first suggested to represent disjunctions in the form of a line connecting two iconic symbols. But in a formal system, the introduction of such symbols requires the definition of transformation rules on diagrams.

$^{ii}$ The analogical/digital distinction also relies on a notion of specificity for Dretske [11]. For him, every signal transmitting information necessarily carries this information under two aspects: an analogical form and a digital form. The analogical form always contains an additional specificity relative to the information properly conveyed by the digital form.

$^{iii}$ The omnipresence of representation of the same type designating the same object is thus observed in human language, where references to an object can be spread out everywhere in a document, so that information is not « localized » (quoted from [13]). For P&M, this additional character is the one required to give diagrammatic systems the closure under constraints property, when combined with iconicity and a constraint and systematic homomorphism.

$^{iv}$ Let us take the example of Ann, Gaston, and Isabel, who are represented as « ordered » in the diagrammatic case. A minimal difference, but an essential one, between the two types of representations is the following:
  (I) left-of (a, g) & left-of (g, i)
and (II) ordered ([ a, g, i]) (or just [a, g, i] )

There is an additional syntactical complexity for (II) which prevents its meaning, contrary to that of (I), from being described as a function of one argument of its predicate's meaning. Indeed, you can easily assign a meaning to the semantic equation: 〚left-of ( a, g) 〛= 〚left-of〛(〚a 〛, 〚g 〛), while you cannot write anything else but: 〚 ordered ([a, g, i]) 〛 = 〚ordered〛 ( 〚 [a, g, i] 〛 ), which implies giving meaning, *at the meta-language level*, to a *configuration* of terms (the list figuring between simple square brackets). Therefore, the semantic descriptive meta-language must *offer possibilities of syntactical structuring of data similar to the ones figuring in the representation language*, because it will sometimes be necessary to assign them a meaning. This is not to say that all syntactical nuances of the representational system must reflected in the interpretation system, because not all iconic representation features are interpreted in a diagrammatic representation (think to the use of marked features in mathematical figure to derive geometrical proofs). Nevertheless, it shows that semantic compositionality relies on syntactic considerations.

[v] Note that in the context of graphical interfaces, several notion of focus are required at a very low level (in the graphic server itself), in order to link the keyboard (and/or events on the pointer of the mouse) to a particular window. The development of graphical interfaces (and networks) has introduced considerable changes in the previous programming framework. (1) There are other sources of input than letters (at least, mouse inputs), and other sorts of output (graphics, sound). (2) The input/output data are of distinct nature, but they may be link together in the system (as the mouse and the screen). (3) The sharing of input/output devices by several programs adds some additional complexity to the emerging framework.